\documentclass{article}

\usepackage{times}
\usepackage{graphicx} 
\usepackage{subfigure}

\usepackage{natbib}

\usepackage{algorithm}
\usepackage{algorithmic}

\usepackage[accepted]{icml2016arxiv}

\usepackage{tikz}
\usetikzlibrary{arrows,shapes,backgrounds,patterns,fadings,decorations.pathreplacing,decorations.pathmorphing}
\tikzset{>=stealth'}
\usetikzlibrary{arrows}
\usepackage{amsmath}
\usepackage{amssymb}
\usepackage{pgfplots} 

\newcommand{\N}{\mathcal{N}}
\newcommand{\C}{{\mathbb C}}
\newcommand{\E}{{\mathbb E}}
\newcommand{\V}{{\mathbb V}}
\newcommand{\pno}[1]{#1_{t|t-1}}   
\newcommand{\uno}[1]{#1_{t|t}}     
\newcommand{\unot}[1]{\tilde #1_{t|t}}     
\newcommand{\pne}[1]{#1_{t+1|t}}   
\newcommand{\now}[1]{#1_t}
\newcommand{\nowt}[1]{\tilde #1_t}
\newcommand{\new}[1]{#1_{t+1}}

\newcommand{\newT}[1]{#1_{T}}
\newcommand{\unotMbm}{\uno{\mu}^{\tilde\bm}}
\newcommand{\unotSbm}{\uno{\Sigma}^{\tilde\bm}}
\definecolor{lgrey}{rgb}{0.8,0.8,0.8}
\newcommand{\argmin}[1]{\underset{#1}{\operatorname{arg}\operatorname{min}}\;}
\newcommand{\inv}{^{-1}}
\newcommand{\nnn}{\nonumber \\}
\newcommand{\nn}{\nonumber}
\newcommand{\fx}{f} 
\newcommand{\fb}{f} 
\newcommand{\bm}{m} 
\newcommand{\BM}{M} 
\newcommand{\der}{\text{d}} 
\newcommand{\expcost}{\mathcal{E}}
\newcommand{\uw}{\text{vec}} 
\newcommand{\newlinetimesdist}{\hspace{0.3cm}\times\;}

\newcommand{\policyparams}{\psi} 
\newcommand{\pendulumangle}{\theta} %
\newcommand{\gplinear}{\phi} 

\newcommand{\note}[1]{}        

\allowdisplaybreaks

\begin{document}

\twocolumn[
\icmltitle{Data-Efficient Reinforcement Learning in Continuous-State POMDPs}

\icmlauthor{Rowan McAllister}{rtm26@cam.ac.uk}
\icmlauthor{Carl Rasmussen}{cer54@cam.ac.uk}
\icmladdress{Department of Engineering, University of Cambridge, Cambridge, CB2 1PZ}

\icmlkeywords{Reinforcement Learning, Data-Efficiency, Filtering}

\vskip 0.3in
]

\begin{abstract}
%


We present a data-efficient reinforcement learning
algorithm
resistant to observation noise. %
Our method extends the highly data-efficient PILCO algorithm \cite{pilco}
into partially observed Markov decision processes (POMDPs)
by considering the filtering process during policy evaluation. 
PILCO conducts policy search,
evaluating each policy by first predicting
an analytic distribution of possible system
trajectories. 
We additionally
predict trajectories w.r.t.\ a filtering process, 
achieving significantly higher performance than
combining a filter
with a policy optimised by the original (unfiltered) framework.
Our test setup is the cartpole swing-up task with sensor noise,
which involves nonlinear dynamics and requires nonlinear control.

\end{abstract}

\section{Introduction}

\note{Context?}
Real world control systems rely on imperfect sensors
to control processes where poor performance often causes real world expense.
Learning to control such a system thus entails:
1) an inability to know the state of the system with certainty
and, 2) penalties for data-inefficiency.
%
%

\subsection{Data Efficiency}
\note{What does the community have (data-efficiency)?}
Most reinforcement learning (RL) methods are data-intensive, requiring much system interaction before learning good policies.
For systems prone to wear and tear, or expensive to operate, data-efficiency is critical.
Model-based RL methods learn models of the unknown system dynamics.
They are generally more data-efficient than model-free RL
because they
1) generalise local dynamics knowledge, and
2) allow local value backups to propagate globally through state-action space.
Unfortunately, a common problem in model-based RL is model bias.
Model bias typically arises when
predictions are based on a single model
selected from a large plausible set,
and then assuming the model is correct with certainty.
An example is using the maximum a posteriori (MAP) model. 
%
Basing predictions off a single model leaves an RL method susceptible to model error.
Any single model is quite possibly the wrong model,
being
just one of the many plausible
explanations of what generated the observed data. 
And the less data observed, the greater the number of plausible dynamics models.
When optimising data-efficiency, 
the agent constantly learns and acts in the low data regime
where the set of plausible models is vast,
exacerbating model bias effects.
This regime undermines traditional trajectory-based control approaches
which assume model-correctness,
such as model predictive control or iterative linear quadratic regulators. 
%
Unless model-based RL algorithms consider the complete set of plausible dynamics,
they will succumb to model-bias,
counteracting the data-efficiency benefits of using a model.

PILCO is a model-based RL algorithm which achieved unprecedented data-efficiency in
learning to control the cartpole swing-up problem
whilst only scaling linearly with horizon \cite{pilco}.
The key to PILCO's success is its 
\textit{probabilistic} dynamics model,
which makes predictions by marginalising over the complete set of
plausible dynamics functions. 
By additionally propagating
model uncertainty throughout trajectory prediction
PILCO avoids model bias.
%
As a result,
PILCO
more likely 
collects data in promising areas of the state space.

\note{Other cites?}

\subsection{Sensor Noise}
\note{More context - how much does noise hurt us?}
The reality of imperfect noisy sensors 
impairs
the control of dynamical systems. 
Such problems can be framed mathematically by
partially observable Markov decision processes (POMDPs).
Solving a POMDP is more complex than its fully observable counterpart, the MDP \cite{pomdp}.
A common approximation to small-noise POMDP problems
is to ignore noise. This assumes full observability
by learning and planning in observation space rather than latent state space.
However, such approximations break down under larger noise levels. 
For example, consider the cartpole system (Figure~\ref{fig:cartpole}).
Stabilising the pendulum upright requires a controller
with a large gain associated with
the pendulum angle.
This enables the cart to move quickly under the pendulum's centre of gravity
in response to slight angle variations.
When incorrectly modelled as a MDP
noise associated with reading the pendulum's angle is injected directly into the policy.
The noise is then
\textit{amplified} by the high gain,
which produces large variation in controller output,
quickly destabilising the system.

\note{What the community have (filtering)?}
The negative effects of sensor noise can be
mitigated using an observation model and filtering.
To \textit{filter} a sequence of sensory outputs is
to maintain a belief posterior distribution
over the latent system state
conditioned on the complete history of previous actions and observations.
Implementing a filter is straightforward
when the system dynamics are \textit{known} and \textit{linear},
referred to as Kalman filtering. 
For nonlinear systems, 
the extended Kalman filter (EKF) is often adequate,
as long as the dynamics are \textit{locally linear},
meaning approximately linear within the region covered by the belief distribution. 
Otherwise, the EKF's first order Taylor expansion approximation breaks down.
Greater nonlinearities usually warrant the unscented Kalman filter (UKF)
or particle methods~\cite{bayesFilters,particleMethods}.
The UKF uses a deterministic sampling technique to estimate moments.
However, if moments can be computed analytically and exactly,
moment-matching methods are preferred.
Moment-matching using distributions from the exponential family (e.g.\ Gaussians)
is equivalent to optimising the Kullback-Leibler divergence $\text{KL}(p||q)$
between the true distribution $p$ and an approximate
distribution $q$.
In such cases, moment-matching is less susceptible to model bias than the EKF 
due to its conservative predictions~\cite{deisenroth2013}.

\subsection{Related Work}
\note{We want...unfortunately the literature does not provide...}
Unfortunately, the literature does
not provide a method that is both data efficient and resistant to noise
when dynamics are \textit{unknown} and \textit{locally nonlinear}.
\note{PILCO fails...}
The original PILCO, which assumes full state observability, fails under moderate sensor noise.
One proposed solution is to filter observations during policy execution \cite{deisenroth2013}.
Filtering during execution does indeed improve performance,
which we demonstrate later.
However, without also predicting system trajectories w.r.t.\ the filtering process,
the above method merely optimises policies for unfiltered control, not for filtered control.
The mismatch between unfiltered-prediction and filtered-execution
restricts PILCO's ability to take full advantage of filtering. 
\citeauthor{dallaire2009} \yrcite{dallaire2009} optimise a policy using a 
more realistic filtered-prediction.
However,
the method neglects model uncertainty by only
using the MAP model.  
Unlike the method of \citeauthor{deisenroth2013} \yrcite{deisenroth2013},
\citeauthor{dallaire2009}'s work \yrcite{dallaire2009}
is therefore highly susceptible to model error,
hampering data-efficiency.

\note{What have we done to fix this?}
We propose the best of both worlds
by extending PILCO from MDPs to POMDPs using full probabilistic predictions w.r.t.\ a filtered process.
We predict using closed loop filtered control precisely because
we execute closed loop filtered control.
The resulting policies are thus optimised for the specific case in which they are used.
Doing so,
our method retains the same data-efficiency properties of PILCO
whilst more resistant to observation noise than PILCO.
To evaluate our method, we use the benchmark cartpole swing-up task with noisy sensors.
We show realistic and probabilistic prediction (to consider uncertainty)
helps our method outperform the aforementioned methods.

\note{Format?}
This paper proceeds by summarising the PILCO framework in greater detail (Section~\ref{sec:pilco}),
which we modify and extend for application to POMDPs (Section~\ref{sec:bayespilco}).
We then compare our method with
the aforementioned methods in the cartpole swing-up problem
(Section~\ref{sec:experiments}),
discussing each method's predicted and empirical performance
(Section~\ref{sec:result-and-analysis}).


\section{The PILCO Algorithm}\label{sec:pilco}

\newcommand{\linedefpolicy}{1}
\newcommand{\lineinitpolicy}{2}
\newcommand{\lineexecute}{4}
\newcommand{\linetrain}{5}
\newcommand{\linepredict}{6}
\newcommand{\lineevaluate}{7}
\newcommand{\lineoptimise}{8}

PILCO is a model-based policy-search RL algorithm.
It applies to continuous-state, continuous-action, continuous-observation and
discrete-time control tasks.
A probabilistic dynamics model is used to predict one-step system dynamics
(from one timestep to the next).
This allows PILCO to probabilistically predict multi-step system trajectories
over arbitrary time horizons $T$,
by repeatedly using the predictive dynamics model's output at one timestep,
as the (uncertain) input
in the following timestep.
For tractability PILCO uses moment-matching to keep the latent state distribution Gaussian.
The result is an analytic distribution of system trajectories,
approximated 
as a joint Gaussian distribution over $T$ states. 
The policy is evaluated as the expected total cost of the trajectories.
Next, the policy is improved using local gradient-based optimisation,
searching over policy-parameter space. 
A distinct advantage of moment-matched prediction for policy search
instead of particle methods
is smoother policy gradients and less local optima~\cite{mchutchon2014}.
Finally, the policy is executed, generating new data
to re-train the dynamics model.
The whole process then repeats until policy convergence.

For the remainder of this section
we discuss, step by step, PILCO summarised by Algorithm~\ref{algo:pilco}.
We first define a policy $\pi$ as a
parametric function (Algorithm~\ref{algo:pilco}, line~\linedefpolicy)
and initialise the policy parameters $\policyparams$ randomly (line~\lineinitpolicy)
since we begin without any data.



\subsection{System Execution}
With a policy now defined, PILCO is ready to \textit{execute} the system (Algorithm~\ref{algo:pilco}, line~\lineexecute).
Let the latent state of the system at time $t$ be $\now{x}\in\mathbb{R}^D$,
which is noisily observed as $\now{z}=\now{x}+\now{\epsilon}$,
where $\now{\epsilon}\stackrel{iid}{\sim}\N(0,\Sigma^\epsilon)$.
The policy $\pi$, parameterised by $\policyparams$,
takes observation $\now{z}$ as input,
and outputs a control action $\now{u}=\pi(\now{z},\policyparams)\in\mathbb{R}^F$.
Applying action $\now{u}$ to the dynamical system in state $\now{x}$,
results in a new system state $\new{x}$. 
This completes the description of system execution,
resulting in a single system-trajectory up until horizon $T$.

\subsection{Learning Dynamics}
To model and learn the unknown dynamics (Algorithm~\ref{algo:pilco}, line~\linetrain),
any probabilistic model flexible enough to capture the complexity of the dynamics can be used.
Bayesian nonparametric models
are particularly suitable because of their resistance to both overfitting and underfitting respectively.
Overfitting otherwise leads to model bias,
and underfitting limits the complexity of the system this method can learn to control.
In a nonparametric model
no prior dynamics knowledge is required, not even knowledge of
\textit{how complex} the unknown dynamics might be
since the model's complexity can grow with the available data.
PILCO chooses to place a Gaussian process (GP) prior on the
latent dynamics function $\fx$.
The training inputs are state-action pairs:
\begin{equation}
\nowt{x} \doteq
\left[\!\begin{array}{c}\now{x}\\ \now{u}\end{array}\!\right] \in \mathbb{R}^{D+F},
\end{equation}
and targets are noisy observations of resultant states\footnote{
The original PILCO GP targets are relative changes in state.}, $\new{z}$.
The covariance function is a square exponential,
\begin{eqnarray}
k(\tilde x_i,\tilde x_j)\;=\;\sigma_f^2\exp\big(-\tfrac{1}{2}(\tilde x_i-\tilde x_j)^\top\Lambda\inv(\tilde x_i-\tilde x_j)\big),
\end{eqnarray}
with length scales $\Lambda=\text{diag}([l_1^2,...,l_{D+F}^2])$,
and signal variance $\sigma_f^2$.
We also use a linear mean function\footnote{
The original PILCO GP uses a zero mean function.}
$\gplinear^\top \tilde x$.
and use the Direct method \cite{mchutchon2014} to train the GP and estimate the noise $\Sigma^\epsilon$
(since the observations are generated from a latent time series).

\subsection{System Prediction}
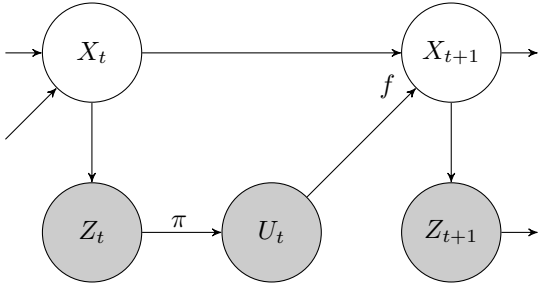
\begin{figure}[t]
\centering
\begin{tikzpicture}[->,>=stealth',scale=1, transform shape]
\node [matrix,matrix anchor=mid, column sep=30pt, row sep=30pt, ampersand
replacement=\&,nodes={circle,draw,minimum size=1.32cm}] {
\node                       (x0)  {$\now{X}$}; \&
                                               \&
\node                       (x1)  {$\new{X}$}; \\
\node[fill=lgrey]           (y0)  {$\now{Z}$}; \&
\node[fill=lgrey]           (u)   {$\now{U}$}; \&
\node[fill=lgrey]           (y1)  {$\new{Z}$}; \\
};
\draw [->] (x0) to (y0)  ;
\draw [->] (y0) to (u) node [left, xshift=-1cm, yshift=0.15cm] {$\pi$} ;
\draw [->] (u) to (x1) ;
\draw [->] (x0) to (x1) node [yshift=-0.45cm, xshift=-0.85cm] {$\fx$} ;
\draw [->] (x1) to (y1)  ;
\draw [->] (x1)  -- ++(1.15,0) ;
\draw [->] (y1)  -- ++(1.15,0) ;
\draw [->] (x0)++(-1.15,0) -- (x0) ;
\draw [->] (x0)++(-1.15,-1.15) -- (x0) ;
\end{tikzpicture}
\caption{\textbf{The original (unfiltered) PILCO, as a directed probabilistic graphical model.}
The latent system $\now{X}$ is observed noisily as $\now{Z}$
which is inputted directly into policy function $\pi$ to decide action $\now{U}$.
Finally, the latent system the will evolve to $\new{X}$,
according to the unknown, nonlinear dynamics function $\fx$ of the previous state $\now{X}$
and action $\now{U}$.
}
\label{fig:ctrlnf} 
\end{figure}

In contrast to executions, PILCO also \textit{predicts} analytic distributions of system trajectories
(Algorithm~\ref{algo:pilco}, line~\linepredict).
It does this offline, between the online system executions, for policy evaluation.
Predicted control is identical to executed control
except each aforementioned quantity is instead now a random variable,
distinguished with capitals:
$\now{X}$, $\now{Z}$, $\now{U}$, $\nowt{X}$ and $\new{X}$,
all approximated as jointly Gaussian.
%
These variables interact
both in execution and prediction
according to Figure~\ref{fig:ctrlnf}.
To predict $\new{X}$ now that 
$\nowt{X}$ is uncertain PILCO uses the iterated law of expectation and variance:
\begin{eqnarray}
p(\new{X}|\nowt{X})&\!\!=\!\!&\N(\new{\mu}^x,\new{\Sigma}^x),\\
\new{\mu}^x&\!\!=\!\!&\E_{\tilde X}[\E_\fx[\fx(\nowt{X})]],\\
\new{\Sigma}^x&\!\!=\!\!&\V_{\tilde X}[\E_\fx[\fx(\nowt{X})]] + \E_{\tilde X}[\V_\fx[\fx(\nowt{X})]].
\end{eqnarray}
After a one-step prediction from $X_0$ to $X_1$,
PILCO repeats the process from $X_1$ to $X_2$, and up to $\newT{X}$,
resulting in a multi-step prediction
whose joint we refer to as a distribution over system trajectories.

\subsection{Policy Evaluation}
To evaluate a policy (or more accurately, a set of policy parameters $\policyparams$),
PILCO applies a cost function to the marginal state distribution at each timestep:
(Algorithm~\ref{algo:pilco}, line~\lineevaluate):
\begin{equation}
 J(\policyparams)=\sum_{t=0}^{T} \gamma^t \now{\expcost}, \quad\quad \now{\expcost}=\E_{X}[\text{cost}(\now{X})|\policyparams].\label{eq:totalcost}
\end{equation}

\subsection{Policy Improvement}
The policy is optimised using the analytic gradients of Eq.~\ref{eq:totalcost}.
A BFGS optimisation method 
searches for the set of policy parameters $\policyparams$
that minimise the total cost $J(\policyparams)$
using gradients information $\der J/\der\policyparams$
(Algorithm~\ref{algo:pilco}, line~\lineoptimise).
To compute $\der J/\der\policyparams$
we require derivatives $\der \now{\expcost} / \der \policyparams$
at each time $t$ to chain together,
and thus $\der p(\now{S}) / \der \policyparams$,
detailed in PILCO \cite{pilco}.

\begin{algorithm}[tb]
\begin{algorithmic}[1]
   \STATE \textit{Define} policy's functional form:
   $\pi: \now{z} \times \policyparams \rightarrow \now{u}$.
   \STATE \textit{Initialise} policy parameters $\policyparams$ randomly.
   \REPEAT
   \STATE \textit{Execute} system, record data.  
   \STATE \textit{Learn} dynamics model.  
   \STATE \textit{Predict} system trajectories from $p(X_0)$ to $p(X_T)$. 
   \STATE \textit{Evaluate} policy: 
   \item[]\quad\quad $J(\policyparams) = \sum_{t=0}^{T} \gamma^t \E_{X} [\text{cost}(X_t) | \policyparams]$.
   \STATE \textit{Optimise} policy: 
   \item[]\quad\quad $\policyparams \leftarrow \argmin{\policyparams} J(\policyparams)$.
   \UNTIL{policy parameters $\policyparams$ converge}
\end{algorithmic}
\caption{PILCO}
\label{algo:pilco}
\end{algorithm}

\section{PILCO Extended with Bayesian Filtering}\label{sec:bayespilco}

\begin{figure}[t]
\centering
\begin{tikzpicture}[->,>=stealth',scale=1, transform shape]
\node [matrix,matrix anchor=mid, column sep=30pt, row sep=30pt, ampersand
replacement=\&,nodes={circle,draw,minimum size=1.32cm}] {
\node                       (x0)  {$\now{X}$}; \&
                                               \&
\node                       (x1)  {$\new{X}$}; \\
\node[fill=lgrey]           (y0)  {$\now{Z}$}; \&
\node[fill=lgrey]           (u)   {$\now{U}$}; \&
\node[fill=lgrey]           (y1)  {$\new{Z}$}; \\
\node[fill=lgrey]           (b00) {$\pno{B}$}; \&
\node[fill=lgrey]           (b10) {$\uno{B}$}; \&
\node[fill=lgrey]           (b11) {$\pne{B}$}; \\
};
\draw [->] (b00) to (b10) ;
\draw [->] (x0) to (y0)  ;
\draw [->] (y0) to (b10) ;
\draw [->] (b10) to (u) node [left, xshift=0.05cm, yshift=-1.2cm] {$\pi$} ;
\draw [->] (b10) to (b11)  node [yshift=0.4cm, xshift=-0.9cm]  {$\fb$} ;
\draw [->] (u) to (b11) ;
\draw [->] (u) to (x1) ;
\draw [->] (x0) to (x1) node [yshift=-0.45cm, xshift=-0.85cm] {$\fx$} ;
\draw [->] (x1) to (y1)  ;
\draw [->] (x1)  -- ++(1.15,0) ;
\draw [->] (y1)-- ++(1.15,-1.15) ;
\draw [->] (b11) -- ++(1.15,0) ;
\draw [->] (x0)++(-1.15,0) -- (x0) ;
\draw [->] (x0)++(-1.15,-1.15) -- (x0) ;
\draw [->] (b00)++(-1.15,0) -- (b00) ;
\draw [->] (b00)++(-1.15,1.15) -- (b00) ;
\end{tikzpicture}
\caption{\textbf{Our method (PILCO extended with Bayesian filtering),
as a directed probabilistic graphical model.}
The latent system (top row) interacts with the agent's belief (bottom row)
via a series of
observations and action decisions (middle row).
At each timestep the latent system $\now{X}$ is observed noisily as $\now{Z}$.
The prior belief $\pno{B}$
(whose dual subscript means belief of the latent physical state at time $t$
given all observations up until time $t-1$ inclusive)
is combined with observation $\now{Z}$
resulting in posterior belief $\uno{B}$ (the update step).
Then, the mean posterior belief $\E[\uno{B}]$
is 
inputted into policy function $\pi$ to decide action $\now{U}$.
Finally, the next timestep's prior belief $\pne{B}$ is predicted using dynamics model $\fb$
(the predict step).
}
\label{fig:ctrlbf} 
\end{figure}
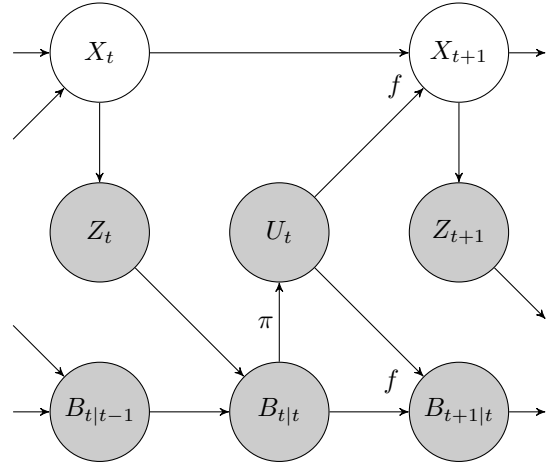

In this section we describe the novel aspects of our method.
Our method uses the same high-level algorithm as PILCO (Algorithm~\ref{algo:pilco}).
However, we modify\footnote{
We implement our method by modifying the PILCO source code from: \tt{http://mlg.eng.cam.ac.uk/pilco/}.} two subroutines
to extend PILCO into POMDPs.
First, we filter observations during system execution (Algorithm~\ref{algo:pilco}, line~\lineexecute)
discussed in Section~\ref{sec:filtered-execution}.
Second, we predict system trajectories w.r.t.\ the filtering process (line~\linepredict),
discussed in Section~\ref{sec:filtered-prediction}.
Filtering maintains a belief distribution of the latent system state.
The belief 
is conditioned on, not just the recent observation,
but all previous actions and observations (Figure~\ref{fig:ctrlbf}).
The extra conditioning provides a less noisy input for the policy:
the belief-mean instead of the raw observation $\now{z}$.

We continue PILCO's distinction between
\textit{executing} the system (resulting in a single real system trajectory)
and \textit{predicting} an analytic distribution of multiple possible system trajectories.
As before, during execution 
the system reads specific observations
and decides specific actions. 
Under probabilistic prediction,
both observations and actions are instead random variables with distributions.
Our method additionally maintains an internal belief state $b$ by filtering observations during execution.
The belief is a random variable, distributed as $b\sim\N(m,V)$.
Consequently, during system prediction 
we consider a distribution over multiple possible belief states, 
i.e.\ a distribution over random variables, 
which we specify with a hierarchical-distribution.

\subsection{Filtered-System Execution}\label{sec:filtered-execution}

When an actual filter is applied, it starts with three pieces of information:
$\pno{\bm}$, $\pno{V}$ and a noisy observation of the system
$\now{z}$.
%
The filtering `update step' combines prior belief $\pno{b}\sim\N(\pno{m},\pno{V})$
with observational likelihood $p(\now{x})=\N(\now{z},\Sigma^\epsilon)$
to yield posterior belief $\uno{b}$:
\begin{eqnarray}
\uno{b}
&\;\sim\;&{\cal N}(\uno{\bm},\uno{V}), \\
\uno{\bm}&\;=\;&W_\bm\pno{\bm}+W_z\now{z},\\
\uno{V}&\;=\;&W_\bm\pno{V},
\end{eqnarray}
with weight matrices $W_\bm=\Sigma^\epsilon(\pno{V}+\Sigma^\epsilon)^{-1}$ and
$W_z=\pno{V}(\pno{V}+\Sigma^\epsilon)^{-1}$.
%
The policy $\pi$
is instead applied to
updated belief-mean $\uno{\bm}$
(a smoother and better-informed signal than $\now{z}$)
to decide action $\now{u}$,
\begin{equation}
\now{u}\;=\;\pi(\uno{\bm},\policyparams).
\end{equation}
Thus, the joint distribution over the
updated (random) belief and the (non-random) action is
\begin{eqnarray}
\unot{b}&\doteq&
\left[\!\begin{array}{c}\uno{b}\\\now{u}\end{array}\!\right],\\
&\sim&{\cal N}\left(\unot{\bm}\doteq\left[\!\begin{array}{c}\uno{\bm}\\\now{u}\end{array}\!\right],\;
\unot{V}\doteq\left[\!\begin{array}{cc}\uno{V}&0\\0&0\end{array}
\!\right]\right). \nn
\end{eqnarray}
Finally, the filtering `prediction step' computes $p(\pne{b})$
as approximately the output of dynamics model $\fb$ with uncertain input
$\unot{b}$.
The output distribution $p(\fb(\unot{b}))$ is non-Gaussian and intractable,
yet has analytically solvable moments \cite{pilco}.
We approximate the distribution $p(\pne{b})$ as Gaussian using moment-matching for tractability:
\begin{eqnarray}
\pne{b}&\;\sim\;&\N(\pne{\bm},\pne{V}),\label{eq:pneB}\\
\pne{\bm}^a&\;=\;&\E_{\unot{b}}[\fb^a(\unot{b})],\label{eq:pneBM}\\
\pne{V}^{ab}&\;=\;&\C_{\unot{b}}[\fb^a(\unot{b}),\;\fb^b(\unot{b})],\label{eq:pnezV}
\end{eqnarray}
where
$\pne{\bm}^a$ and $\pne{V}^{ab}$ are derived in Appendix~\ref{sec:app-instantiation-prediction}.
The
process then repeats
using the predictive belief (Eq.~\ref{eq:pneB}-\ref{eq:pnezV})
as the prior belief in the following timestep.
This completes the specification of the system in execution.

\subsection{Filtered-System Prediction}\label{sec:filtered-prediction}

In \emph{system prediction},
we compute the probabilistic behaviour of the
filtered system via an analytic distribution of possible beliefs.
A distribution over beliefs $b$ is in principle a distribution over
its parameters $\bm$ and $V$.
To distinguish $\bm$ and $b$ as now being
\textit{random} and \textit{hierarchically-random}
respectively,
we capitalise them: $\BM$ and $B$. 
As an approximation we are going
to assume that the distribution on the variance $\pno{V}$
is a delta
function (i.e.\ some fixed value, for a given timestep).
Restricting $\BM$ to being Gaussian distributed then
we begin system prediction with the joint:
\begin{equation}
\left[\!\!\begin{array}{c}\pno{\BM}\\\now{Z}\end{array}\!\!\right]
\sim\N\left(\left[\!\!\begin{array}{c}\pno{\mu}^\bm \\ \now{\mu}^x\end{array}\!\!\right]\!,\!
\left[\!\!\begin{array}{cc}\pno{\Sigma}^\bm&0\\0&\now{\Sigma}^z\end{array}\!\!\right]\right),
\end{equation}
where $\now{\Sigma}^z=\now{\Sigma}^x+\Sigma^\epsilon$.

\paragraph{Remark:} We now pause for a moment
to reflect on the full model of the filtered system (Figure~\ref{fig:ctrlbf}).
The model is slightly more general than a POMDP.
For instance,
it can predict the consequences of a mismatch between
the latent state $\now{X}$ and the agent's belief $\pno{B}$.
Such a feature is perhaps not very interesting, however,
since from the agent's point of view, the latent state is unknown.
All the agent's knowledge of the latent state is
summarised by its belief.
A special case of our framework that reduces exactly to a POMDP is as follows.
We constrain the latent state distribution to be a `flattened'
version of the hierarchically-distributed belief:
$\now{X}\sim\N(\now{\mu}^x,\now{\Sigma}^x)$, where $\now{\mu}^x=\pno{\mu}^\bm$
and $\now{\Sigma}^x=\pno{\Sigma}^\bm+\pno{V}$.
Additionally we use identical dynamics function $f$
for both latent and belief dynamics.
Doing so, the Markov system state in Figure~\ref{fig:ctrlbf}
reduces from $\{\now{X},\pno{B}\}$ to just $\{\pno{B}\}$.
I.e.\ 
predicting the next latent state $\new{X}\sim\N(\new{\mu}^x,\new{\Sigma}^x)$
is conditionally independent of $\now{X}$ given $\pno{B}$.
In such case, $p(\new{X})$ does not need to be explicitly computed,
since an analogous relationship holds true:
$\new{\mu}^x=\pne{\mu}^\bm$
and $\new{\Sigma}^x=\pne{\Sigma}^\bm+\pne{V}$.
We will use this special POMDP case throughout the rest of this paper
for system prediction.
Multi-step prediction, which requires a Markov state from one timestep to the next,
now simply predicts from one set of beliefs to the next.
This is the belief-MDP interpretation of POMDPs \cite{beliefMDPs}.

Moving on, the updated belief posterior is also Gaussian,
\begin{equation}
\uno{\BM}\;\sim\;\N\left(\uno{\mu}^\bm,\uno{\Sigma}^\bm\right),
\end{equation}
where $\uno{\mu}^\bm = \pno{\mu}^\bm$ and
$\uno{\Sigma}^\bm=
W_m\pno{\Sigma}^mW_m^\top + W_z\now{\Sigma}^zW_z^\top$.
The policy now has a random input $\uno{\BM}$, thus
the control output must also be random
(even though we use a deterministic policy function):
\begin{equation}
 \now{U}\;=\;\pi(\uno{\BM},\policyparams),
\end{equation}
which we implement by overloading the policy function:
\begin{equation}
(\now{\mu}^u,\now{\Sigma}^u,\now{C}^{\bm u})
\;=\;\pi(\uno{\mu}^\bm,\uno{\Sigma}^\bm,\policyparams),
\end{equation}
where $\now{\mu}^u$ is the output mean, $\now{\Sigma}^u$ the output variance and
$\now{C}^{\bm u}$ input-output covariance with premultiplied
inverse input variance, $\now{C}^{\bm u} \doteq (\uno{\Sigma}^\bm)^{-1}\C_{\BM}[\uno{\BM},\now{U}]$. 
Making a moment-matched approximation yields a joint Gaussian:
\begin{eqnarray}
\!\!\!\!&\!\!\!\!\!\unot{\BM}\doteq\left[\!\!\begin{array}{c}\uno{\BM}\\ \now{U}\end{array}\!\!\right]\label{eq:postA}\\
\!\!\!\!&\!\!\sim\!\N
\!\left(\!\unotMbm\doteq\!\!\left[\!\!\!\begin{array}{c}\uno{\mu}^\bm\\ \now{\mu}^u\end{array}\!\!\!\right]\!,
\unotSbm\doteq\!\left[\!\!\!\!\begin{array}{cc}\uno{\Sigma}^\bm&\!\!\!\!\uno{\Sigma}^\bm \now{C}^{\bm u}\\
(\now{C}^{\bm u})^\top\uno{\Sigma}^\bm&\!\!\!\!\now{\Sigma}^u\end{array}\!\!\!\right]\!\right)\label{eq:postB}
\end{eqnarray}
Finally, we probabilistically predict
1) the belief-mean distribution $p(\pne{\BM})$ and
2) the expected belief-variance $\pne{\bar V}=\E[\pne{V}]$,
both detailed in Appendix~\ref{sec:app-simulation-prediction}.
We have now discussed the one-step prediction of the filtered system,
from $\pno{B}$ to $\pne{B}$ 
Using this process repeatedly, from initial belief $B_{0|0}$
we predict forwards to $B_{1|0}$, then to $B_{2|1}$ etc., up to $B_{T|T-1}$. 

\subsection{Policy Evaluation and Improvement}
To evaluate a policy we again
apply the cost function (Eq.~\ref{eq:totalcost})
to the multi-step prediction (Section~\ref{sec:filtered-prediction}).
Note the marginal distribution of each latent state $\now{X}$ at time $t$
is related to the belief $\pno{B}$ by:
\begin{eqnarray}
\now{X}&\sim&\N(\pno{\mu}^\bm,\pno{\Sigma}^m+\pno{V}) \quad\forall\;t,
\end{eqnarray}
where the belief is hierarchically distributed:
$\pno{B}\sim\N(\pno{M},\pno{V})\sim\N(\N(\pno{\mu}^\bm,\pno{\Sigma}^m),\pno{V})$.
The policy is again optimised using the analytic gradients of Eq.~\ref{eq:totalcost},
except now we consider how filtering affects the gradients of $\now{X}$.
%
Let $\uw(\cdot)$ be the `unwrap operator'
that reshapes a matrix into a vector.
We can define a Markov filtered-system from the belief's parameters:
$\now{S}
=[\pno{\BM}^\top,\;\uw(\pno{V})^\top]^\top$.
To predict system evolution, the state distribution is defined (further details in Appendix~\ref{sec:app-gradients}):
\begin{equation}
p(\now{S})
\!\sim\!\N\!\left(\!\now{\mu}^s\!=\!\begin{bmatrix} \pno{\mu}^\bm \\ \uw(\pne{V}) \end{bmatrix}\!, 
\now{\Sigma}^s\!=\!\begin{bmatrix} \pno{\Sigma}^\bm &\!\!\!0 \\0 &\!\!\!0\end{bmatrix}\!\right).
\end{equation}

\section{Experiments}\label{sec:experiments}

\begin{figure}[h] 
\centering
\begin{tikzpicture}
\def\xcart {2.5}     
\def\ycart {0.25}  
\def\wheelR {0.1} 
\def\cartW  {0.8} 
\def\cartH {0.4}  
\def\ang {22}     
\def\xtipR {-1}   
\def\ytipR {2.25} 
\def\goallen {0.2} 
\def\pendlen {2.427} 
  \draw[draw=black,ultra thick] (-1,0) -- (4,0);
 \draw[draw=blue,fill=blue] (\xcart-\cartW/2,\wheelR) rectangle (\xcart+\cartW/2,\wheelR+\cartH);
 \draw[draw=black,fill=black] (\xcart-\cartW/2,\wheelR) circle (\wheelR cm);
 \draw[draw=black,fill=black] (\xcart+\cartW/2,\wheelR) circle (\wheelR cm);
 \node[anchor=east] at (\xcart-\cartW+0.3,\ycart) {$m_c$};
 \draw[draw=red, line width = 3pt] (\xcart,\ycart) -- (\xcart+\xtipR,\ycart+\ytipR)node[anchor= east] {$m_p$}  node[anchor= east, midway]  {$l$};
 \draw[draw=black, line width = 3pt] (\xcart+\xtipR,\ycart+\ytipR) circle (0.04cm);
 \draw[draw=black, dashed] (\xcart,\ycart-1.5) -- (\xcart,\ycart+2.5);
 \draw[draw=black] (0,-.5) -- (0,.5);
 \draw [->] ([shift=(90:2)]\xcart,\ycart)  arc (90:90+\ang:2) node[xshift=0.35cm, yshift=0.35cm] {$\pendulumangle$};
 \draw [|->|,draw=black] (0,-1) -- (\xcart,-1) node[fill=white, midway] {$x_c$};
 \draw[|->|,draw=black] (\xcart+\xtipR+2.5,\ycart) -- (\xcart+\xtipR+2.5,\ycart+\ytipR) node[fill=white, midway]  {$y_p$};
 \draw[|->|,draw=black] (0,\ycart+\ytipR+1) -- (\xcart+ \xtipR,\ycart+\ytipR+1) node[fill=white, midway]  {$x_p$};
 \draw[draw=black, line width = 2pt] (0,\pendlen+\ycart-\goallen) -- (0,\pendlen+\ycart+\goallen);
 \draw[draw=black, line width = 2pt] (-\goallen,\pendlen+\ycart) -- (\goallen,\pendlen+\ycart);
 \draw[|->|,draw=black] (-0.5,\ycart) -- (-0.5,\pendlen+\ycart) node[fill=white, midway] {$l$};
 \draw[very thick,draw=green,->]  (\xcart,\ycart) -- (\xcart +1,\ycart) node[anchor=south]  {$u$};
\end{tikzpicture}
\caption{\textbf{The cartpole swing-up task.}
A pendulum of length $l$ is attached to a cart by a frictionless pivot.
The cart has mass $m_c$ and position $x_c$.
The pendulum's endpoint has mass $m_p$ and position $(x_p,y_p)$, with angle $\pendulumangle$ from vertical.
The system begins with cart at position $x_c=0$ and pendulum hanging down: $\pendulumangle=\pi$.
The goal is to accelerate the cart by applying horizontal force $u_t$ at each timestep $t$
to invert then stabilise the pendulum's endpoint at the goal (black cross),
i.e.\ to maintain $x_c=0$ and $\pendulumangle=0$.
}
\label{fig:cartpole} 
\end{figure}
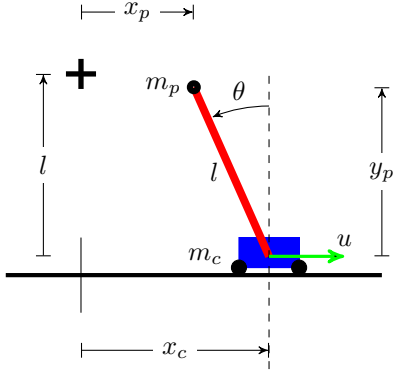

We test our algorithm on the cartpole swing-up problem (Figure~\ref{fig:cartpole}),
a benchmark for comparing controllers of nonlinear dynamical systems.
We experiment using a physics simulator by solving the differential equations of the system.
The pendulum begins each episode hanging downwards
with the goal of swinging it up and stabilising it.

We now describe our test system.
The cart has mass $m_c = 0.5\text{kg}$.
A zero-order hold controller applies horizontal forces to the cart within range $[-10,10]\text{N}$.
The controller / policy is a radial basis function with 100 centroids.
Friction resists the cart's motion with damping coefficient $b=0.1\text{Ns/m}$.
Connected to the cart is a pole of length $l=0.2\text{m}$
and mass $m_p = 0.5\text{kg}$ located at its endpoint,
which swings due to gravity's acceleration $g=9.82\text{m/s}^2$.
An inexpensive camera observes the system.
Frame rates of \$10 webcams are typically 30Hz at maximum resolution,
thus the time discretisation is $\Delta t = 1/30 s$.
The state $x$ comprises the cart position, pendulum angle, and
their time derivatives $x=[x_c,\pendulumangle,\dot x_c, \dot\pendulumangle]^\top$.
The cartpole's motion is described with the differential equation:
\vspace{-1mm}
\begin{eqnarray}
\dot x=\left[\!\!\!\begin{array}{c}\dot{x_c}\\ \dot{\pendulumangle}\\
\displaystyle \frac{-2m_pl \dot{\pendulumangle}^2s+3m_pgsc+4u-4b\dot{x_c}}
{4(m_c+m_p)-3m_pc^2}\\[1em] \displaystyle
\frac{-3m_pl \dot{\pendulumangle}^2sc+6(m_c+m_p)gs+6(u-b\dot{x_c})c}
{4l(m_c+m_p)-3m_plc^2}\end{array}\!\!\!\right],
\end{eqnarray}
using shorthand $s=\sin\pendulumangle$ and $c=\cos\pendulumangle$.
Both the 
initial latent state and initial belief
are i.i.d.: 
$X_0,B_{0|0}\stackrel{iid}{\sim}\N(M_{0|0},V_{0|0})$
where $M_{0|0} \sim \delta([0,\pi,0,0]^\top)$ and $V_{0|0}^{\frac{1}{2}} = \text{diag}([0.2\text{m},0.2\text{rad},0.2\text{m/s},0.2\text{rad/s}])$.
The camera's noise standard deviation is:
$(\Sigma^\epsilon)^\frac{1}{2} =
\text{diag}([0.03\text{m}, 0.03\text{rad}, \frac{0.03}{\Delta t}\text{m/s}, \frac{0.03}{\Delta t}\text{rad/s}])$, 
noting $0.03\text{rad}\approx 1.7^\circ$. 
We use the $\frac{0.03}{\Delta t}$ terms since using a camera we cannot observe velocities directly
but can estimate with finite differences,
and thus the observation error is dependent on the observation error of the positions.
Each episode has a two second time horizon (60 timesteps). 
The cost function 
we impose is
$1-\exp\left(-\frac{1}{2}d^2/\sigma_c^2\right)$
where $\sigma_c = 0.25m$
and $d^2$ is the squared Euclidean distance between the pendulum's end point $(x_p,y_p)$
and its goal $(0,l)$.
I.e. $d^2 = x_p^2 + (l-y_p)^2 = (x_c-l\sin\pendulumangle)^2 + (l-l\cos\pendulumangle)^2$.

We compare four algorithms:
1) PILCO \cite{pilco} as a baseline (unfiltered execution, and unfiltered full-prediction);
2) the method by \cite{dallaire2009} (filtered execution, and filtered MAP-prediction);
3) the method by \cite{deisenroth2013} (filtered execution, and unfiltered full-prediction);
and lastly
4) our method (filtered execution, and filtered full-prediction).
For clear comparison
we opted for a tightly controlled experiment.
We control for data and dynamics models,
i.e.\ each algorithm has access to the exact same data and exact same dynamics model.
The reason is to
eliminate variance in performance caused by different algorithms choosing different actions.
We generate a single dataset by running the baseline PILCO algorithm
for 11 episodes (totalling 22 seconds of system interaction). 
The independent variables of our experiment are
1) the method of system prediction and
2) the method of system execution.
We then optimise each policy from the same initialisation
using their respective prediction methods.
Finally, we measure and compare their performances
in both prediction and execution.

\section{Results and Analysis}\label{sec:result-and-analysis}

We now compare algorithm performance, both predictive (Figure~\ref{fig:simulation-results})
and from empirical execution (Figure~\ref{fig:empirical-results}).

\subsection{Predictive Performance}
First, we analyse predictive costs per timestep (Figure~\ref{fig:simulation-results}).
Since predictions are probabilistic,
the costs have distributions, 
with the exception of \citeauthor{dallaire2009} \yrcite{dallaire2009} which predicts MAP trajectories
and therefore has deterministic cost.
Even though we plot distributed costs,
policies are optimised w.r.t.\ expected total cost only.
Using the same dynamics,
the different prediction methods optimise different policies
(with the exception of \cite{pilco} and \cite{deisenroth2013},
whose prediction methods are identical).
During the first 10 timesteps, we note 
identical performance with maximum cost
due to the non-zero time required physically swing the pendulum up
near the goal.
Performances thereafter diverge.
Since we predict w.r.t.\ a filtering process,
less noise is predicted to be injected into the policy,
and the optimiser can thus afford higher gain parameters w.r.t.\ the pole at balance point.
If we linearise our policy around the goal point,
our policy has a gain of -81.7N/rad w.r.t.\ pendulum angle,
a larger-magnitude than both Deisenroth method gains of -39.1N/rad
(negative values refer to \textit{left} forces in Figure~\ref{fig:cartpole}).
Being afforded higher gains
our policy is more reactive and more likely to catch a falling pendulum.
Finally,
we note \citeauthor{dallaire2009} \yrcite{dallaire2009} predict very high performance.
Without balancing the costs across multiple possible trajectories,
the method instead optimises a sequence of deterministic states to near perfection.

\vspace{-2mm}
\subsection{Empirical Performance}
We now compare the predictive results
against the empirical results, using 100 executions of each algorithm (Figure~\ref{fig:empirical-results}).
First, we notice a stark difference between predictive and executed performances from \cite{dallaire2009},
due to neglecting model uncertainty, suffering model bias.
In contrast, the other methods consider uncertainty and have relatively unbiased predictions,
judging by the similarity between predictive-vs-empirical performances.
Deisenroth's methods, which differ only in execution,
illustrate that filtering during execution-only can be better than no filtering at all.
However, the real benefit comes when
the policy is evaluated from multi-step predictions of a filtered system.
Opposed to \citeauthor{deisenroth2013}'s method \yrcite{deisenroth2013},
our method's predictions reflect reality closer
because we both predict and execute system trajectories
using closed loop filtering control. 

\vspace{-2mm}
\section{Conclusion and Future Work}\label{sec:conclusions}
%
%

\newcommand{\xmins}{0}
\newcommand{\horizon}{60} 
\newcommand{\ymins}{0}
\newcommand{\ymaxs}{1}
\newcommand{\xticks}{0,5,10,15,20,25,30,35,40,45,50,55,60}
\newcommand{\xlabels}{Timestep}
\newcommand{\ylabels}{Cost}
\newcommand{\legends}{legend entries={
(NFexe | NFsim),
Dallaire (BFexe | BFMAPsim),
Deisenroth (BFexe | NFsim),
CtrlBF (BFexe | BFsim)},}
\newcommand{\legendstyle}{legend style={at={(0.96,0.96)},anchor=north east}}


\pgfplotscreateplotcyclelist{rowan}{%
blue,every mark/.append style={fill=blue!80!black},mark=*\\%
red,every mark/.append style={fill=red!80!black},mark=square*\\%
brown!60!black,every mark/.append style={fill=brown!80!black},mark=triangle*\\%
black,mark=x\\%
}

\begin{figure}[t]
\vspace{-3mm}
\begin{tikzpicture}
\begin{axis}[
xlabel=\xlabels,
ylabel=\ylabels,
ylabel absolute, ylabel style={yshift=-0.15cm},
xtick={\xticks},
xmin=\xmins,
xmax=\horizon,
ymin=\ymins,
ymax=\ymaxs,
legend entries={Deisenroth \yrcite{pilco},Dallaire \yrcite{dallaire2009},Deisenroth \yrcite{deisenroth2013},Our Method},
\legendstyle,
legend cell align=left,
cycle list name=rowan,
]
\addplot+[
error bars/.cd,
y dir=both,y explicit,
]
table[x=x,y=y,y error=error]
{
x y error
0 0.999823 0.000155
1 0.999827 0.000150
2 0.999832 0.000142
3 0.999846 0.000127
4 0.999836 0.000136
5 0.999740 0.000221
6 0.999356 0.000572
7 0.998250 0.001573
8 0.996866 0.002805
9 0.996615 0.003152
10 0.997790 0.002215
11 0.999109 0.000974
12 0.999737 0.000297
13 0.999883 0.000140
14 0.999676 0.000598
15 0.996638 0.006727
16 0.975856 0.038894
17 0.909621 0.118964
18 0.793397 0.229103
19 0.667309 0.312287
20 0.575818 0.345038
21 0.518318 0.350500
22 0.478913 0.346400
23 0.451211 0.340671
24 0.433531 0.336600
25 0.423815 0.334779
26 0.420211 0.334727
27 0.420953 0.335741
28 0.424309 0.337254
29 0.428875 0.338906
30 0.433752 0.340500
31 0.438415 0.341932
32 0.442536 0.343144
33 0.445885 0.344096
34 0.448309 0.344767
35 0.449723 0.345148
36 0.450102 0.345240
37 0.449478 0.345054
38 0.447926 0.344605
39 0.445558 0.343915
40 0.442513 0.343011
41 0.438951 0.341929
42 0.435045 0.340709
43 0.430974 0.339402
44 0.426918 0.338060
45 0.423052 0.336741
46 0.419536 0.335503
47 0.416508 0.334399
48 0.414080 0.333473
49 0.412331 0.332755
50 0.411301 0.332262
51 0.410998 0.331993
52 0.411402 0.331931
53 0.412473 0.332047
54 0.414178 0.332304
55 0.416508 0.332665
56 0.419520 0.333092
57 0.423381 0.333553
58 0.428433 0.334024
59 0.435295 0.334471
60 0.444997 0.334829
};
\addplot+[
error bars/.cd,
y dir=both,y explicit,
]
table[x=x,y=y,y error=error]
{
x y error
0 0.999665 0.000000
1 0.999665 0.000000
2 0.999664 0.000000
3 0.999635 0.000000
4 0.999460 0.000000
5 0.999084 0.000000
6 0.998943 0.000000
7 0.999274 0.000000
8 0.999624 0.000000
9 0.999711 0.000000
10 0.999156 0.000000
11 0.992031 0.000000
12 0.942226 0.000000
13 0.799472 0.000000
14 0.590125 0.000000
15 0.389524 0.000000
16 0.230935 0.000000
17 0.113921 0.000000
18 0.040760 0.000000
19 0.018846 0.000000
20 0.027261 0.000000
21 0.012930 0.000000
22 0.003097 0.000000
23 0.001813 0.000000
24 0.003007 0.000000
25 0.003576 0.000000
26 0.003493 0.000000
27 0.002572 0.000000
28 0.002061 0.000000
29 0.001251 0.000000
30 0.000751 0.000000
31 0.000503 0.000000
32 0.000469 0.000000
33 0.000309 0.000000
34 0.000307 0.000000
35 0.000497 0.000000
36 0.000309 0.000000
37 0.000065 0.000000
38 0.000022 0.000000
39 0.000206 0.000000
40 0.000533 0.000000
41 0.000733 0.000000
42 0.000903 0.000000
43 0.000811 0.000000
44 0.000721 0.000000
45 0.000471 0.000000
46 0.000310 0.000000
47 0.000129 0.000000
48 0.000051 0.000000
49 0.000004 0.000000
50 0.000001 0.000000
51 0.000016 0.000000
52 0.000022 0.000000
53 0.000043 0.000000
54 0.000039 0.000000
55 0.000060 0.000000
56 0.000053 0.000000
57 0.000081 0.000000
58 0.000077 0.000000
59 0.000116 0.000000
60 0.000114 0.000000
};
\addplot+[
error bars/.cd,
y dir=both,y explicit,
]
table[x=x,y=y,y error=error]
{
x y error
0 0.999823 0.000155
1 0.999827 0.000150
2 0.999832 0.000142
3 0.999846 0.000127
4 0.999836 0.000136
5 0.999740 0.000221
6 0.999356 0.000572
7 0.998250 0.001573
8 0.996866 0.002805
9 0.996615 0.003152
10 0.997790 0.002215
11 0.999109 0.000974
12 0.999737 0.000297
13 0.999883 0.000140
14 0.999676 0.000598
15 0.996638 0.006727
16 0.975856 0.038894
17 0.909621 0.118964
18 0.793397 0.229103
19 0.667309 0.312287
20 0.575818 0.345038
21 0.518318 0.350500
22 0.478913 0.346400
23 0.451211 0.340671
24 0.433531 0.336600
25 0.423815 0.334779
26 0.420211 0.334727
27 0.420953 0.335741
28 0.424309 0.337254
29 0.428875 0.338906
30 0.433752 0.340500
31 0.438415 0.341932
32 0.442536 0.343144
33 0.445885 0.344096
34 0.448309 0.344767
35 0.449723 0.345148
36 0.450102 0.345240
37 0.449478 0.345054
38 0.447926 0.344605
39 0.445558 0.343915
40 0.442513 0.343011
41 0.438951 0.341929
42 0.435045 0.340709
43 0.430974 0.339402
44 0.426918 0.338060
45 0.423052 0.336741
46 0.419536 0.335503
47 0.416508 0.334399
48 0.414080 0.333473
49 0.412331 0.332755
50 0.411301 0.332262
51 0.410998 0.331993
52 0.411402 0.331931
53 0.412473 0.332047
54 0.414178 0.332304
55 0.416508 0.332665
56 0.419520 0.333092
57 0.423381 0.333553
58 0.428433 0.334024
59 0.435295 0.334471
60 0.444997 0.334829
};
\addplot+[
error bars/.cd,
y dir=both,y explicit,
]
table[x=x,y=y,y error=error]
{
x y error
0 0.999823 0.000155
1 0.999827 0.000150
2 0.999833 0.000142
3 0.999846 0.000127
4 0.999832 0.000140
5 0.999711 0.000245
6 0.999233 0.000671
7 0.997848 0.001886
8 0.996135 0.003376
9 0.995729 0.003818
10 0.997115 0.002680
11 0.998836 0.001149
12 0.999690 0.000327
13 0.999892 0.000127
14 0.999772 0.000380
15 0.997542 0.004660
16 0.979608 0.033694
17 0.916769 0.115772
18 0.798489 0.232677
19 0.660504 0.317801
20 0.551887 0.344801
21 0.474032 0.341390
22 0.411023 0.325682
23 0.357098 0.304626
24 0.313520 0.283384
25 0.280472 0.265417
26 0.256868 0.251875
27 0.240539 0.242015
28 0.228459 0.234131
29 0.218333 0.226938
30 0.209063 0.219895
31 0.200042 0.212730
32 0.190891 0.205277
33 0.181474 0.197502
34 0.171884 0.189499
35 0.162370 0.181451
36 0.153233 0.173574
37 0.144755 0.166083
38 0.137143 0.159166
39 0.130515 0.152962
40 0.124905 0.147563
41 0.120278 0.143002
42 0.116558 0.139266
43 0.113643 0.136301
44 0.111423 0.134025
45 0.109790 0.132342
46 0.108644 0.131156
47 0.107898 0.130376
48 0.107477 0.129928
49 0.107318 0.129745
50 0.107369 0.129779
51 0.107586 0.129986
52 0.107937 0.130335
53 0.108396 0.130799
54 0.108941 0.131359
55 0.109558 0.131996
56 0.110234 0.132697
57 0.110960 0.133450
58 0.111724 0.134242
59 0.112516 0.135061
60 0.113323 0.135893
};
\end{axis}
\end{tikzpicture}
\caption{\textbf{Predictive costs per timestep.}
The error bars show $\pm1$ standard deviation.
Each algorithm has access to the same data set
(generated by baseline Deisenroth \yrcite{pilco}) and the same dynamics model.
Algorithms differ in their multi-step prediction methods
(except Deisenroth's algorithms whose predictions thus overlap).
}
\label{fig:simulation-results} 
\end{figure}
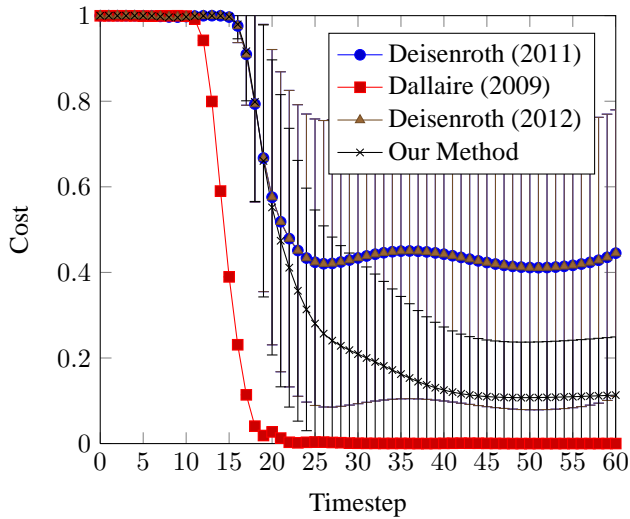

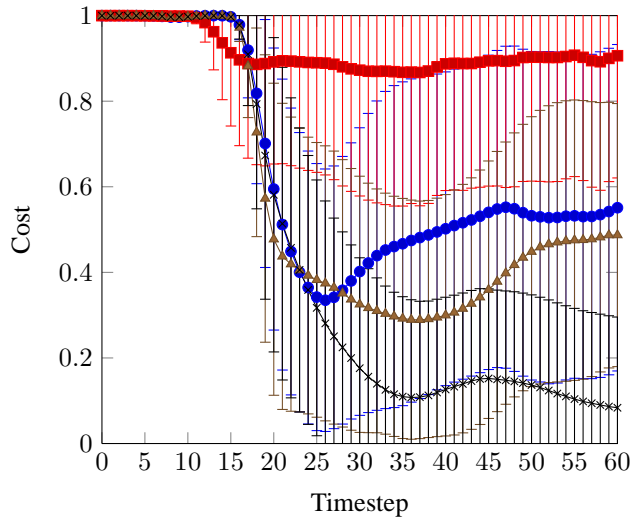
\begin{figure}[t]
\vspace{-3mm}
\begin{tikzpicture}
\begin{axis}[
xlabel=\xlabels,
ylabel=\ylabels,
ylabel absolute, ylabel style={yshift=-0.15cm},
xtick={\xticks},
xmin=\xmins,
xmax=\horizon,
ymin=\ymins,
ymax=\ymaxs,
\legendstyle,
legend cell align=left,
cycle list name=rowan,
]
\addplot+[
error bars/.cd,
y dir=both,y explicit,
]
table[x=x,y=y,y error=error]
{
x y error
0 0.999825 0.000143
1 0.999826 0.000144
2 0.999824 0.000148
3 0.999840 0.000132
4 0.999833 0.000137
5 0.999732 0.000222
6 0.999334 0.000571
7 0.998143 0.001680
8 0.996499 0.003311
9 0.996275 0.003623
10 0.997558 0.002423
11 0.999056 0.000987
12 0.999744 0.000288
13 0.999897 0.000121
14 0.999716 0.000472
15 0.997151 0.005223
16 0.978623 0.034958
17 0.919340 0.111550
18 0.818274 0.211081
19 0.701096 0.289706
20 0.594720 0.329549
21 0.511670 0.339923
22 0.448809 0.335153
23 0.400393 0.326285
24 0.364594 0.318912
25 0.342228 0.312019
26 0.334893 0.306699
27 0.341698 0.306939
28 0.358190 0.312670
29 0.380001 0.322682
30 0.401867 0.336018
31 0.421848 0.349512
32 0.439094 0.361455
33 0.452214 0.370568
34 0.460246 0.376195
35 0.466842 0.378345
36 0.474457 0.377711
37 0.481276 0.376039
38 0.487487 0.373065
39 0.494507 0.368989
40 0.501865 0.366595
41 0.509305 0.365765
42 0.515878 0.365537
43 0.522456 0.364840
44 0.531084 0.365474
45 0.540220 0.367286
46 0.548004 0.370497
47 0.552191 0.376214
48 0.548647 0.380126
49 0.539456 0.381984
50 0.532279 0.383863
51 0.529728 0.385926
52 0.527736 0.388108
53 0.528228 0.386814
54 0.531204 0.382670
55 0.532152 0.379647
56 0.530198 0.378579
57 0.530582 0.378954
58 0.534945 0.380466
59 0.542119 0.381522
60 0.551022 0.381415
};
\addplot+[
error bars/.cd,
y dir=both,y explicit,
]
table[x=x,y=y,y error=error]
{
x y error
0 0.999826 0.000153
1 0.999835 0.000141
2 0.999842 0.000130
3 0.999812 0.000161
4 0.999714 0.000301
5 0.999537 0.000578
6 0.999314 0.001031
7 0.999074 0.001849
8 0.998935 0.002495
9 0.998950 0.002923
10 0.998581 0.003447
11 0.995129 0.014237
12 0.983573 0.045513
13 0.962273 0.089074
14 0.936380 0.132907
15 0.913112 0.171147
16 0.896629 0.200576
17 0.888126 0.221301
18 0.886036 0.234435
19 0.887881 0.239044
20 0.891748 0.238551
21 0.894313 0.240432
22 0.893710 0.244455
23 0.891843 0.248319
24 0.890495 0.252287
25 0.889803 0.256148
26 0.889123 0.261458
27 0.886291 0.269711
28 0.880303 0.278434
29 0.875539 0.286300
30 0.872479 0.294021
31 0.869975 0.301149
32 0.869667 0.307298
33 0.870129 0.311777
34 0.868889 0.313956
35 0.867468 0.312524
36 0.867993 0.307155
37 0.867085 0.311542
38 0.870524 0.309461
39 0.879883 0.300186
40 0.887252 0.296092
41 0.888583 0.296230
42 0.887947 0.296304
43 0.888050 0.291305
44 0.891107 0.292199
45 0.894285 0.292711
46 0.894764 0.292928
47 0.893332 0.294980
48 0.894954 0.294937
49 0.902073 0.289712
50 0.902877 0.289720
51 0.902687 0.290116
52 0.902168 0.290752
53 0.902197 0.289961
54 0.904357 0.285763
55 0.907389 0.283607
56 0.902522 0.288012
57 0.894760 0.295721
58 0.892180 0.300427
59 0.900044 0.287975
60 0.906186 0.285544
};
\addplot+[
error bars/.cd,
y dir=both,y explicit,
]
table[x=x,y=y,y error=error]
{
x y error
0 0.999826 0.000155
1 0.999828 0.000152
2 0.999832 0.000146
3 0.999852 0.000123
4 0.999848 0.000130
5 0.999750 0.000222
6 0.999349 0.000596
7 0.998106 0.001723
8 0.996298 0.003286
9 0.995895 0.003585
10 0.997145 0.002431
11 0.998816 0.000996
12 0.999654 0.000302
13 0.999863 0.000129
14 0.999696 0.000367
15 0.996907 0.004039
16 0.972168 0.032593
17 0.881971 0.120297
18 0.727233 0.243755
19 0.572574 0.335825
20 0.477382 0.364601
21 0.437354 0.357690
22 0.419146 0.343671
23 0.405295 0.331356
24 0.392585 0.323302
25 0.382463 0.320966
26 0.374337 0.321340
27 0.364398 0.318990
28 0.350977 0.312414
29 0.336877 0.305765
30 0.325324 0.299335
31 0.316906 0.291487
32 0.310065 0.284870
33 0.303491 0.281585
34 0.297206 0.280650
35 0.292254 0.280515
36 0.289448 0.279400
37 0.289059 0.277550
38 0.291000 0.276984
39 0.294777 0.278630
40 0.299759 0.282024
41 0.306320 0.286320
42 0.315265 0.290314
43 0.327038 0.293852
44 0.341840 0.297315
45 0.359443 0.300838
46 0.378519 0.303991
47 0.398177 0.306413
48 0.417466 0.308439
49 0.434607 0.311291
50 0.448328 0.315395
51 0.458789 0.319717
52 0.465923 0.323413
53 0.470251 0.326329
54 0.473202 0.327723
55 0.475209 0.327592
56 0.475881 0.325519
57 0.477634 0.321169
58 0.482075 0.314750
59 0.487000 0.310337
60 0.487111 0.307362
};
\addplot+[
error bars/.cd,
y dir=both,y explicit,
]
table[x=x,y=y,y error=error]
{
x y error
0 0.999848 0.000178
1 0.999848 0.000173
2 0.999851 0.000162
3 0.999870 0.000134
4 0.999864 0.000140
5 0.999762 0.000245
6 0.999380 0.000650
7 0.998256 0.001852
8 0.996644 0.003548
9 0.996292 0.003917
10 0.997376 0.002796
11 0.998922 0.001172
12 0.999709 0.000312
13 0.999863 0.000259
14 0.999568 0.001493
15 0.996989 0.005781
16 0.977657 0.032451
17 0.909582 0.119862
18 0.793178 0.244679
19 0.672879 0.335658
20 0.581303 0.367245
21 0.513817 0.365315
22 0.457281 0.350521
23 0.405697 0.332024
24 0.358907 0.314160
25 0.317085 0.298858
26 0.280732 0.286122
27 0.250543 0.276274
28 0.224720 0.269888
29 0.199725 0.264423
30 0.176009 0.258031
31 0.156212 0.252326
32 0.139801 0.247310
33 0.125480 0.242527
34 0.115266 0.238268
35 0.109663 0.233474
36 0.107617 0.228402
37 0.108440 0.223978
38 0.112459 0.220528
39 0.119339 0.217918
40 0.126318 0.215520
41 0.132916 0.213740
42 0.139331 0.212448
43 0.146013 0.211700
44 0.150999 0.210864
45 0.152125 0.209338
46 0.150095 0.208457
47 0.147816 0.208757
48 0.145137 0.209522
49 0.141355 0.210223
50 0.136846 0.210548
51 0.131035 0.210815
52 0.123886 0.211704
53 0.116751 0.212000
54 0.110176 0.211826
55 0.103903 0.211934
56 0.098312 0.212003
57 0.094192 0.212008
58 0.090206 0.212105
59 0.086442 0.212156
60 0.083606 0.212024
};
\end{axis}
\end{tikzpicture}
\caption{
\textbf{Empirical costs per timestep}.
We generate empirical cost distributions from 100 executions per algorithm.
Error bars show $\pm1$ standard deviation.
The plot colours and shapes correspond to the legend in Figure~\ref{fig:simulation-results}.
}
\label{fig:empirical-results} 
\end{figure}

In this paper,
we extended the original PILCO algorithm \cite{pilco} to filter observations,
both during system execution and multi-step probabilistic prediction required for policy evaluation.
The extended framework enables learning in \textit{partially-observed} environments (POMDPs)
whilst retaining PILCO's data-efficiency property.
We demonstrated successful application to a benchmark control problem,
the noisily-observed cartpole swing-up.
Our algorithm 
learned a good policy under significant observation noise in less than 30 seconds of system interaction. 
Importantly, our algorithm evaluates policies with predictions that are faithful to reality.
We predict w.r.t.\ closed loop filtered control precisely because we execute closed loop filtered control.

We showed experimentally that \textit{faithful} and \textit{probabilistic}
predictions give greater performance gains than otherwise.
For clear comparison we constrained each algorithm to use the same dynamics dataset
rather than each interacting with the system to generate their own.
If we relaxed this experimental constraint, 
we anticipate our method's performance gains would be greater still.
However, the extra variance in empirical performance
(caused by selection of different data)
means a much larger number of experiments
is required to test if such an \textit{additional} performance gain exists, 
which we plan to do in future work.

Several more challenges remain for future work.
Firstly the assumption of zero variance of the belief-variance could be relaxed.
A relaxation allows distributed trajectories to more accurately
consider belief states having various degrees of certainty (belief-variance).
E.g. system trajectories have larger belief-variance when passing though
data-sparse regions of state-space,
and smaller belief-variance in data-dense regions.
Secondly, the policy could be a function of the full belief distribution (mean and variance)
rather than just the mean. Such flexibility could help the policy make more `cautious' actions when more uncertain about the state.
Thirdly, the framework could be extended to active learning.
Currently, the framework is a passive learner,
greedily optimising the total cost-means
and ignoring cost-variance information which could otherwise better inform exploration,
increasing data-efficiency further.

\appendix
\section{Dynamics Predictions in System-Execution} \label{sec:app-instantiation-prediction} 
\vspace{-2mm}
Here we specify the predictive distribution $p(\pne{b})$,
whose moments are equal to the moments from dynamics model output $\fb$ with uncertain input
$\unot{b}$:
\begin{eqnarray}
\pne{b}&\;\sim\;&\N(\pne{\bm},\pne{V}),\label{eq:pneb}\\
\pne{\bm}^a&\;=\;&\E_{\unot{b}}[\fb^a(\unot{b})]=s_a^2\beta_a^\top q^a\!+\!\gplinear^\top\unot{\bm},\label{eq:pnez}\\ 
C_a&\;=\;&\unot{V}\inv\C_{\unot{b}}[\unot{b},\fb^a(\unot{b})-\gplinear^\top\unot{b}],\nnn
&\;=\;&\;s^2_a(\Lambda_a+\unot{V})^{-1}(\text{x}-\unot{\bm})\beta_aq^a,\label{eq:pneC}\\
\pne{V}^{ab}&\;=\;&\C_{\unot{b}}[\fb^a(\unot{b}),\;\fb^b(\unot{b})],\nnn
&\;=\;&s_a^2s_b^2\big[\beta_a^\top ( Q^{ab}- q^a q^{b\top})\beta_b+\nnn
&&\delta_{ab}\big(s_a^{-2}-\operatorname{tr}((K_a+\Sigma_\varepsilon^a)^{-1} Q^{aa})\big)\big]+\nnn
&& C_a^\top\unot{V} \gplinear_b + \gplinear_a^\top\unot{V} C_b + \gplinear_a^\top\unot{V} \gplinear_b, \label{eq:pneV}\\
 q^a_i&\;=\;&q\big(\text{x}_i,\unot{\bm},\Lambda_a,\unot{V}\big),\label{eq:pneq}\\
 Q^{ab}_{ij}&\;=\;&Q\big(\text{x}_i,\text{x}_j,\Lambda_a,\Lambda_b,0,\unot{\bm},\unot{V}\big).\label{eq:pneQ}
\end{eqnarray}
\vspace{-2mm}
where,
\begin{flalign}
&q(\text{x}_i,\mu,\Lambda,V)\;\doteq\;|\Lambda^{-1}V+I|^{-1/2}\nnn
&\newlinetimesdist\exp\big(-\tfrac{1}{2}(\text{x}_i-\mu)[\Lambda+V]^{-1}(\text{x}_i-\mu)\big),\\
&Q(\text{x}_i,\text{x}_j,\Lambda_a,\Lambda_b,V, \mu, \Sigma)\;\doteq\;|R|^{-1/2}\,\nnn
&\newlinetimesdist q(\text{x}_i,\mu,\Lambda_a,V)\,q(\text{x}_j,\mu,\Lambda_b,V)\nnn
&\newlinetimesdist\exp\big(\tfrac{1}{2}\text{z}_{ij}^\top R^{-1}\Sigma\text{z}_{ij}\big),\\
&R\;=\;\Sigma\big((\Lambda_a+V)^{-1}+(\Lambda_b+V)^{-1}\big)+I,\\
&\text{z}_{ij}\;=\;(\Lambda_a\!+\!V)^{-1}(x_i\!-\!\mu)+(\Lambda_b\!+\!V)^{-1}(x_j\!-\!\mu),\\
&\beta_a\;=\;(K_a+\Sigma^{\epsilon,a})^{-1}(y_a-\gplinear_a^\top\text{x}),
\end{flalign}
and training inputs are $\text{x}$, outputs are $y_a$,
and the GP linear mean function has weight-vector $\gplinear\in\mathbb{R}^D$.
\vspace{-2mm}
\section{Dynamics Predictions in System-Prediction} \label{sec:app-simulation-prediction} 
\vspace{-2mm}
Here we describe the prediction formulae for the random belief state in system-prediction.
We again note, during execution, our belief distribution is specified by certain parameters,
$\uno{b}\sim\N(\uno{\bm},\uno{V})$.
By contrast, during system prediction, 
our belief distribution is specified by an uncertain belief-mean and certain belief-variance:
$\uno{B}\sim\N(\uno{\BM},\uno{V})\sim\N(\N(\uno{\mu}^m,\uno{\Sigma}^m),\uno{\bar V})$,
where we assumed a delta distribution on $V$: $\uw(\uno{V})\sim\N(\uw(\uno{\bar V}),0)$
for mathematical simplicity.
Therefore we conduct GP prediction given hierarchically-uncertain inputs,
giving rise to the various subsection below: 
\vspace{-2mm}
\subsection{Mean of the Belief-Mean}
\vspace{-2mm}
Dynamics prediction uses input
$\unot{\BM}\sim \N(\unotMbm,\unotSbm)$,
which is jointly distributed according to Eq.~\ref{eq:postA}-\ref{eq:postB}.
%
Using the belief-mean $\pne{\bm}^a$ definition (Eq.~\ref{eq:pnez}),
\vspace{-2mm}
\begin{eqnarray}
\pne{\mu}^{\bm,a}&\;=\;&\E_{\unot{\BM}}[\pne{\BM}^a],\nnn
&\;=\;&\int \pne{\BM}^a \N(\unot{\BM}|\unotMbm,\unotSbm)\der\unot{\BM},\nnn
&\;=\;&s_a^2\beta_a^\top \hat q^a,\\
\hat q^a_i&\;=\;&q\Big(x_i,\unotMbm,\Lambda_a,\unotSbm+\unot{V}\Big).
\end{eqnarray}
\vspace{-8mm}
\subsection{Input-Output Covariance}
\vspace{-2mm}
The
expected input-output covariance belief term (Eq.~\ref{eq:pneC})
(also the input-output covariance of the belief-mean) is:
\begin{eqnarray}
\hat C_a
&\;=\;&\unot{V}\inv\E_{\unot{\BM}}[\C_{\uno{B}}[\unot{B},f(\unot{B})-\gplinear_a^\top \unot{\BM}]], \nnn
&\;=\;&(\unotSbm)\inv\C_{\unot{\BM}}[\unot{\BM},\E_{\uno{B}}[f(\unot{B}) - \gplinear_a^\top \unot{\BM}]], \nnn
&\;=\;&s^2_a(\Lambda_a+\unotSbm+\unot{V})^{-1}(\text{x}\!-\!\unotMbm)\beta_a\hat q^a_i.
\end{eqnarray}
\vspace{-8mm}
\subsection{Variance of the Belief-Mean}
\vspace{-2mm}
The variance of randomised belief-mean (Eq~\ref{eq:pnez}) is:
\vspace{-2mm}
\begin{eqnarray}
\pne{\Sigma}^{\bm,ab}&\;=\;&\C_{\unot{\BM}}[\pne{\BM}^a,\;\pne{\BM}^b],\nnn
&\;=\;&\int\!\!\pne{\BM}^a \pne{\BM}^b \N(\unot{\BM}|\unotMbm,\unotSbm)\der\unot{\BM}-\nnn
&&\mu_{\pne{\bm}}^a \mu_{\pne{\bm}}^b,\nnn
&\;=\;&s_a^2s_b^2\beta_a^\top(\hat Q^{ab}-\hat q^a \hat q^{b\top})\beta_b + \nnn
&&\hat C_a^\top\unotSbm \gplinear_b + \gplinear_a^\top\unotSbm \hat C_b + \gplinear_a^\top\unotSbm \gplinear_b, \\
\hat Q_{ij}^{ab}&\;=\;&Q(\text{x}_i,\text{x}_j,\Lambda_a,\Lambda_b,\unot{V},\unotMbm,\unotSbm).
\end{eqnarray}
\vspace{-8mm}
\subsection{Mean of the Belief-Variance}
\vspace{-2mm}
Using the belief-variance $\pne{V}^{ab}$ definition (Eq.~\ref{eq:pneV}),
\vspace{-2mm}
\begin{eqnarray}
\pne{\bar V}^{ab}&\;=\;&\E_{\unot{\BM}}[\pne{V}^{ab}],\nnn
&\;=\;&\int \pne{V}^{ab} \N(\unot{\BM}|\unotMbm,\unotSbm)\der\unot{\BM},\nnn
&\;=\;&s_a^2s_b^2\big[\beta_a^\top (\tilde Q^{ab}-\hat Q^{ab})\beta_b+\nnn
&&\delta_{ab}\big(s_a^{-2}-\operatorname{tr}((K_a+\Sigma_\varepsilon^a)^{-1}\tilde
Q^{aa})\big)\big] + \nnn 
&&\hat C_a^\top\unot{V} \gplinear_b + \gplinear_a^\top\unot{V} \hat C_b + \gplinear_a^\top\unot{V} \gplinear_b, \\
\tilde Q_{ij}^{ab}&\;=\;&Q(\text{x}_i,\text{x}_j,\Lambda_a,\Lambda_b,0,\unotMbm,\unotSbm\!+\!\unot{V}).
\end{eqnarray}

\vspace{-2mm}
\section{Gradients for Policy Improvement} \label{sec:app-gradients} 
\vspace{-2mm}
To compute policy gradient $\der J/\der\policyparams$
we first require $\der \now{\expcost} / \der \policyparams$:
\vspace{-2mm}
\begin{eqnarray}\frac{\der \now{\expcost}}{\der\theta}&\;=\;&
\frac{\der\now{\expcost}}{\der p(\now{S})} \frac{\der p(\now{S})}{\der \theta}, \nnn
&\;=\;& \frac{\der\now{\expcost}}{\partial \now{\mu}^s} \frac{\partial \now{\mu}^s}{\der \theta} +
\frac{\der\now{\expcost}}{\partial \now{\Sigma}^s} \frac{\partial \now{\Sigma}^s}{\der \theta},
\end{eqnarray}
\vspace{-2mm}
and
\begin{eqnarray}
\frac{\der p(\new{S})}{\der\theta} = \frac{\der p(\new{S})}{\der p(\now{S})}
\frac{\der p(\now{S})}{\der\theta} + \frac{\partial p(\new{S})}{\partial\theta}.
\end{eqnarray}

\vspace{-2mm}
Application of the chain rule backwards from the state distribution at the horizon $S_T$,
to $S_t$ at arbitrary time $t$,
is analogous to that detailed in
PILCO \cite{pilco},
where we use $\now{S}$, $\now{\mu}^s$ and $\now{\Sigma}^s$ in the place of
$\now{x}$, $\now{\mu}$ and $\now{\Sigma}$ respectively.

%
%

\bibliography{icml2016}
\bibliographystyle{icml2016}

\end{document}